\PassOptionsToPackage{table}{xcolor}
\documentclass[runningheads]{llncs}
 
\usepackage{eccv}

\usepackage{eccvabbrv}

\usepackage{graphicx}
\usepackage{booktabs}
\usepackage{multirow}
\usepackage{threeparttable}
\usepackage{enumitem}
\usepackage[accsupp]{axessibility}  

\usepackage{hyperref}

\usepackage{orcidlink}

\title{Guide, Think, Act: Interactive Embodied Reasoning in Vision-Language-Action Models}

\titlerunning{Guide, Think, Act}

\author{
Yiran Ling\inst{2,3,7}\orcidlink{0009-0005-3334-1762}$^{*,\ddagger}$ \and
Qing Lian\inst{1}\orcidlink{0000-0002-7608-1084}$^{*}$ \and
Jinghang Li\inst{1,4}\orcidlink{0000-0001-6196-6165} \and
Qing Jiang\inst{3,5}\orcidlink{0009-0003-2067-6411} \and
Tianming Zhang\inst{6}\orcidlink{0009-0003-8923-9498} \and
Xiaoke Jiang\inst{3,6}\orcidlink{0009-0005-9096-5045} \and
Chuanxiu Liu\inst{6}\orcidlink{0009-0008-8566-9737} \and
Jie Liu\inst{2,7}\orcidlink{0000-0001-6209-6886}$^{\dagger}$ \and
Lei Zhang\inst{1,3,6}\orcidlink{0000-0001-6926-0538}$^{\dagger}$
}

\authorrunning{Y. Ling, Q. Lian et al.}

\institute{
 Futian Laboratory \and
 Faculty of Computing, Harbin Institute of Technology \and
 International Digital Economy Academy (IDEA) \and
 School of Robotics, Hunan University \and
 South China University of Technology \and
 Visincept \and
 National Key Laboratory of Smart Farm Technologies and Systems
}

\begin{document}

\maketitle

\begingroup
\footnotesize
\renewcommand{\thefootnote}{}
\footnotetext{\hspace{-1.8em}$^{*}$\ Equal contribution. Emails: \texttt{25b903029@stu.hit.edu.cn}, \texttt{lianqing1997@gmail.com}}
\footnotetext{\hspace{-1.8em}$^{\ddagger}$\ Internship at Futian~Laboratory.\quad $^{\dagger}$~Corresponding~author}
\endgroup

\begin{abstract}
    In this paper, we propose \textbf{GTA-VLA} (\textbf{Guide, Think, Act}), an interactive Vision-Language-Action (VLA) framework that enables spatially steerable embodied reasoning by allowing users to guide robot policies with explicit visual cues. Existing VLA models learn a direct "Sense-to-Act" mapping from multimodal observations to robot actions. While effective within the training distribution, such tightly coupled policies are brittle under out-of-domain (OOD) shifts and difficult to correct when failures occur. Although recent embodied Chain-of-Thought (CoT) approaches expose intermediate reasoning, they still lack a mechanism for incorporating human spatial guidance, limiting their ability to resolve visual ambiguities or recover from mistakes. To address this gap, our framework allows users to optionally guide the policy with spatial priors, such as affordance points, boxes, and traces, which the subsequent reasoning process can directly condition on. Based on these inputs, the model generates a unified spatial-visual Chain-of-Thought that integrates external guidance with internal task planning, aligning human visual intent with autonomous decision-making. For practical deployment, we further couple the reasoning module with a lightweight reactive action head for efficient action execution. Extensive experiments demonstrate the effectiveness of our approach. On the in-domain SimplerEnv WidowX benchmark, our framework achieves a state-of-the-art 81.2\% success rate. Under OOD visual shifts and spatial ambiguities, a single visual interaction substantially improves task success over existing methods, highlighting the value of interactive reasoning for failure recovery in embodied control.
\end{abstract}

\section{Introduction}

The pursuit of robust generalist robotic agents for open-world environments is a central goal of embodied AI.
A major step toward this vision is the emergence of Vision-Language-Action (VLA) models~\cite{rt22023arxiv,kimOpenVLAOpenSourceVisionLanguageAction2024,liCogACTFoundationalVisionLanguageAction2024,teamOctoOpenSourceGeneralist2024,chiDiffusionPolicyVisuomotor2023,black2024pi0,blackP05VisionLanguageActionModel,zheng2025xvla,dengGraspVLAGraspingFoundation2025,cheangGR3TechnicalReport2025,nvidia2025gr00t,chenInternVLAM1SpatiallyGuided2025}, which leverage large pre-trained vision language models to scale robot learning across diverse tasks and embodiments. Despite this progress, most existing VLAs still operate through an implicit direct ``Sense-to-Act'' mapping from multimodal observations to robot actions. While effective within the training distribution, such tightly coupled policies often become brittle under visual and semantic shifts, and provide little transparency when failures occur. When perception fails due to clutter, lighting variation, or unseen objects, humans have no explicit interface to re-ground the robot's attention or provide targeted corrective guidance.

\begin{figure}[t]
    \centering
    \includegraphics[width=1\linewidth]{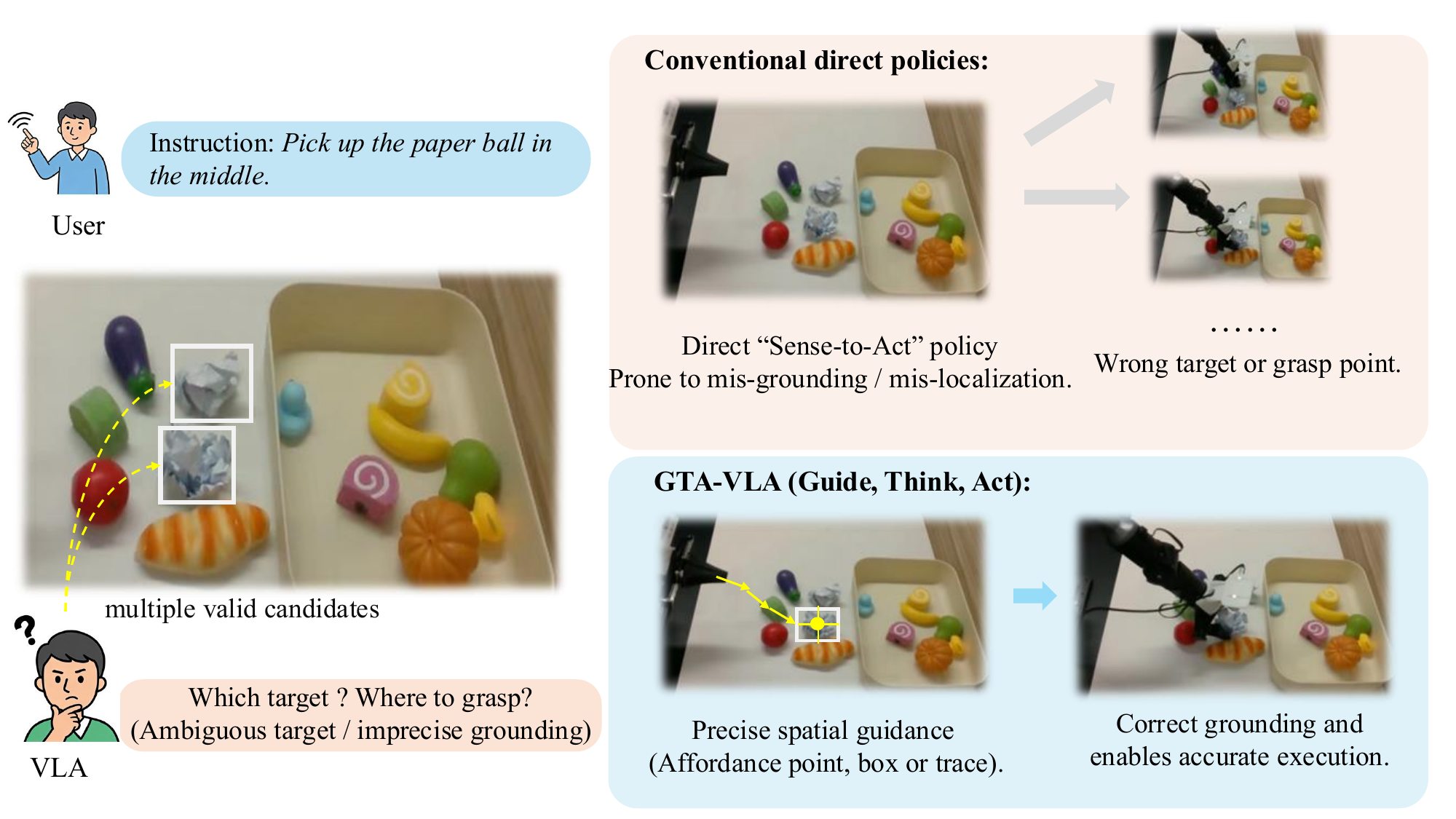}
    \caption{Conventional direct VLA policies can fail under spatial ambiguity or imprecise grounding, since they lack an explicit mechanism for interactive correction. \textbf{GTA-VLA} resolves this by using one-shot spatial guidance (affordance points, boxes, or traces) to correct grounding and enable accurate execution.}
    \label{fig:teaser}
\end{figure}

Recent work has begun to move beyond direct ``Sense-to-Act'' policies through Embodied Chain-of-Thought (CoT) reasoning~\cite{Belkhale2024RTHAH,huangThinkActVisionLanguageActionReasoning2025,Gu2023RTTrajectoryRT,zawalskiRoboticControlEmbodied2025a,tangMindHandPurposeful2025,leeMolmoActActionReasoning2025}, shifting toward a more structured ``Sense, Think, and Act'' paradigm. By explicitly predicting intermediate representations, such as task decomposition, grounding cues, or motion plans, these methods improve interpretability and expose part of the policy's decision-making process. However, in existing systems, the reasoning process remains largely self-contained: although intermediate reasoning is visible, it is still generated from the model's internal belief alone and cannot be easily corrected when that internal grounding is wrong. This limitation is especially pronounced in out-of-domain (OOD) settings, where early mistakes in object grounding, contact localization, or motion targeting can propagate into coherent but incorrect plans. In such cases, human users can often resolve the ambiguity immediately through simple spatial cues, such as pointing to a target, marking a grasp region, or sketching a desired path. Compared with language-only correction, these signals are more precise and more natural for communicating spatial intent, motivating the need for an interaction interface that allows human guidance to directly condition the policy's reasoning process.

To address this gap, we propose \textbf{GTA-VLA}, (\textbf{Guide, Think, Act}), an interactive VLA framework that makes embodied reasoning explicitly steerable under human spatial guidance. The key idea is to treat affordances, boxes, and trajectories not as post-hoc corrections but as optional priors that the model's reasoning process can directly condition on. With or without such guidance, the model produces a unified spatial-visual Chain-of-Thought that integrates external spatial intent with internal task understanding, visual grounding, affordance prediction, and action planning. As a result, the policy remains autonomous by default while becoming naturally correctable when failures or ambiguities arise.

To make this capability trainable at scale, we build an automated data pipeline that synthesizes large-scale interactive annotations from existing robot datasets, without requiring manual collection of human intervention traces. To mitigate the latency of autoregressive reasoning in practical control, we further decompose policy execution into a slow VLM reasoning module and a fast downstream action head. This asynchronous design allows high-level spatial-visual reasoning to run at a lower frequency, while a lightweight action module executes responsive low-level control at a higher frequency. To evaluate interactive embodied reasoning under distribution shift, we introduce \textit{SimplerEnv-Plus}, an extension of SimplerEnv with more challenging OOD conditions, including camera variation, lighting changes, unseen objects, and language perturbations, while also supporting human spatial intervention during execution. Experiments in both simulation and the real world show that our framework improves not only autonomous task performance, but also failure recovery through minimal human interaction.

In summary, our contributions are three-fold:
\begin{enumerate}[nosep,leftmargin=*,topsep=2pt]
    \item We propose GTA-VLA (\textit{Guide, Think, Act}), an interactive VLA framework that unifies explicit human spatial guidance with embodied Chain-of-Thought reasoning, enabling more steerable and interpretable robot policies.
    \item We develop a scalable data generation pipeline for synthesizing interaction-style supervision from existing robot datasets, making guided spatial reasoning trainable at scale.
    \item We introduce \textit{Simpler-Plus}, a benchmark for evaluating interactive embodied policies under OOD conditions. Experiments in simulation and the real world demonstrate strong autonomous performance as well as substantial gains in failure recovery from minimal human intervention.
\end{enumerate}
\section{Related Work}

\noindent{\textbf{Vision-Language-Action Models.}}
Vision-Language-Action (VLA) models learn policies that map visual observations and language instructions to robot actions, and have shown strong generalization across diverse scenes, tasks, and embodiments. Early end-to-end approaches~\cite{rt22023arxiv, kimOpenVLAOpenSourceVisionLanguageAction2024,liCogACTFoundationalVisionLanguageAction2024,teamOctoOpenSourceGeneralist2024,chiDiffusionPolicyVisuomotor2023} demonstrated the effectiveness of scaling imitation learning for real-world embodied manipulation. More recent dual-system architectures~\cite{black2024pi0,blackP05VisionLanguageActionModel,zheng2025xvla,dengGraspVLAGraspingFoundation2025,cheangGR3TechnicalReport2025,nvidia2025gr00t,chenInternVLAM1SpatiallyGuided2025} further improve performance by pairing a vision-language backbone with a dedicated continuous-control module. Despite these advances, current VLA systems still rely heavily on large-scale behavioral data~\cite{o2024open,wu2025robomind,khazatsky2024droid} and predominantly learn implicit action policies from demonstrations. As a result, while they are effective at direct policy execution, they offer limited support for explicit task understanding, interactive correction, and user-guided control when failures or ambiguities arise.

\noindent{\textbf{Visual and Embodied Reasoning and Chain of Thought.}}
Recent work has explored explicit intermediate reasoning as a way to improve the generalization and interpretability of VLA policies. Several approaches~\cite{Belkhale2024RTHAH,huangThinkActVisionLanguageActionReasoning2025,Gu2023RTTrajectoryRT} formulate robotic control as a structured reasoning process rather than direct action prediction. For example, ECoT~\cite{zawalskiRoboticControlEmbodied2025a} and Mind2Hand~\cite{tangMindHandPurposeful2025} introduce embodied Chain-of-Thought (CoT) reasoning to improve high-level planning and policy interpretability, while MolmoAct~\cite{leeMolmoActActionReasoning2025} further grounds intermediate reasoning in explicit spatial representations. A practical challenge in this line of work is that autoregressively generating explicit reasoning tokens can introduce substantial inference latency, which limits responsiveness in manipulation tasks. To improve efficiency, more recent methods have explored compact reasoning representations. For instance, Fast-ThinkAct~\cite{huangFastThinkActEfficientVisionLanguageAction2026} replaces explicit token generation with latent planning states to reduce inference cost. While such designs improve efficiency, they may also reduce the transparency and fine-grained controllability of the reasoning process compared with explicit spatially grounded intermediate representations.

\noindent{\textbf{Interactive Perception and Visual Prompting.}}
Recent work on interactive perception has substantially improved the spatial grounding ability of foundation models. In computer vision, the SAM family~\cite{kirillov2023segany,ravi2024sam2,carionSAM3Segment2025} demonstrates that simple geometric prompts can support strong zero-shot segmentation, while subsequent systems such as T-Rex2~\cite{Jiang2024TRex2TG} and Rex-Omni~\cite{jiang2025detect} extend this paradigm to open-vocabulary and interactive object detection with both text and visual prompts. In parallel, multimodal large language models (MLLMs)~\cite{Chen2023ShikraUM,qwen3vl,seed2025seed1_5vl} have also become increasingly capable of fine-grained visual grounding. Works such as Ferret~\cite{You2023FerretRA,Yang2023SetofMarkPU} and Set-of-Mark (SoM) show that language models can be conditioned on points, boxes, and marks to support precise spatial reference and coordinate-level reasoning.
While these advances have significantly strengthened interactive 2D perception and grounding, extending such capabilities to embodied control remains nontrivial. Most existing visual-prompting systems are designed primarily for pixel-level perception tasks, such as segmentation, detection, or spatial reference, rather than for generating continuous motor actions. As a result, they do not directly address the temporal dynamics, control interfaces, or action generation requirements needed for robotic manipulation.
\section{Methodology}
\begin{figure*}[t]
    \centering
    \includegraphics[width=\textwidth]{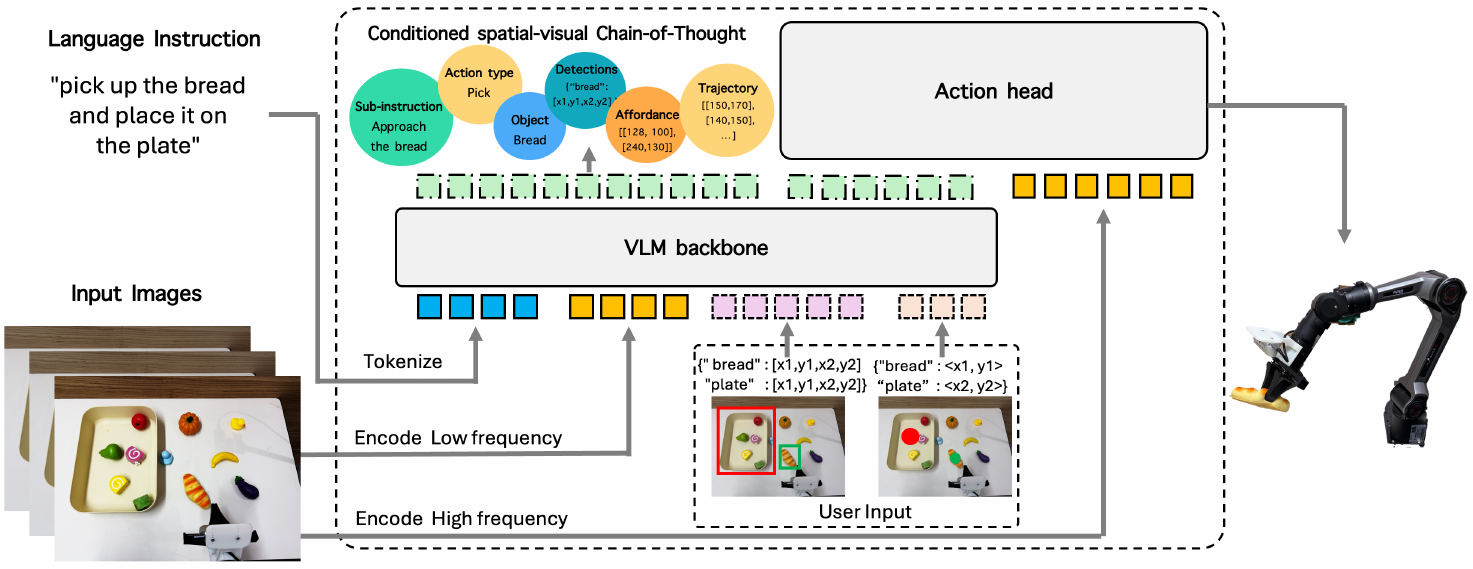}
    \caption{\textbf{Overview of \textbf{GTA-VLA} (\textbf{Guide, Think, Act}).} The framework consists of three stages. \textbf{Guide:} the model receives the primary image, the language instruction, and optional spatial priors (\textit{e.g.}, affordance points, boxes, or traces). \textbf{Think:} the VLM backbone generates a conditioned spatial-visual reasoning sequence and the corresponding latent reasoning states $H_{\text{reasoning}}$. \textbf{Act:} a downstream Flow-Matching action head consumes the latest reasoning states together with high-frequency control observations to produce continuous action chunks. This design decouples slow autoregressive reasoning from fast closed-loop control.}
    \label{fig:method_overview}
\end{figure*}

\subsection{Preliminaries and Overall Architecture}

\noindent\textbf{Preliminaries.}
We formulate robotic manipulation as a conditional sequence modeling problem. At time step $t$, a standard Vision-Language-Action (VLA) policy $\pi$ receives multi-view RGB observations $\mathcal{I}_t = \{I_t^{(v)}\}_{v=1}^{V}$, a natural language instruction $L$, and the robot proprioceptive state $s_t$, and predicts a future action chunk
\begin{equation}
    A_t = [a_t, a_{t+1}, \dots, a_{t+k-1}],
\end{equation}
where $k$ denotes the size of the action chunk. The standard policy is therefore written as
\begin{equation}
    \pi : (\mathcal{I}_t, L, s_t) \rightarrow A_t.
\end{equation}
In our implementation, the multi-view observations consist of a primary external view and a wrist-mounted view.

We extend this formulation by introducing an optional spatial prior $P_{\text{spatial}}$, which provides sparse geometric guidance in image space and may be supplied either by a human user or by an expert annotation pipeline during training. Concretely, $P_{\text{spatial}}$ may take the form of an affordance point, a bounding box, or a trace. The policy is thus extended to
\begin{equation}
    \pi : (\mathcal{I}_t, L, s_t, P_{\text{spatial}}) \rightarrow A_t,
\end{equation}
where $P_{\text{spatial}}$ is optional and may be absent during fully autonomous execution.

\noindent\textbf{Overall Architecture.}
Our framework is built on top of a vision-language backbone and a downstream continuous control module. We use Qwen3-VL-2B~\cite{qwen3vl} as the core VLM backbone due to its strong multimodal understanding and spatial grounding capabilities. Given the augmented input, the backbone first generates a structured spatial-visual reasoning sequence $C$, and we use the hidden states associated with these reasoning tokens as the latent reasoning representation, denoted by $H_{\text{reasoning}}$. In our implementation, the reasoning branch consumes only the primary image, the language instruction, and the optional spatial prior, while proprioceptive inputs and the wrist-view image are introduced only in the downstream fast control branch. These latent reasoning states are then consumed by a downstream action model to generate continuous control actions.

For action generation, we adopt a Flow-Matching action head~\cite{black2024pi0,nvidia2025gr00t,zheng2025xvla}, which models action chunks in continuous space and is well-suited for complex multi-modal action distributions. Concretely, the action branch takes the current control observation together with the latest reasoning states, combining the primary image, the wrist-view image, proprioceptive inputs, and $H_{\text{reasoning}}$ to predict continuous action chunks. In our implementation, actions are primarily parameterized as end-effector poses, although the framework is not restricted to this choice. While our experiments focus on single-arm manipulation, the same formulation can be extended to dual-arm settings by enlarging the action space.

\noindent\textbf{The \textit{Guide-Think-Act} Paradigm.}
Based on this formulation, we decompose policy inference into three stages, as illustrated in Fig.~\ref{fig:method_overview}:
\begin{itemize}
    \item \textbf{Guide (Sec.~\ref{subsec:guide}):} We incorporate optional spatial priors $P_{\text{spatial}}$ into the observation stream, allowing human users to provide sparse geometric guidance alongside the language instruction.
    
    \item \textbf{Think (Sec.~\ref{subsec:think}):} Instead of directly predicting actions from observations, the VLM generates a structured spatial-visual reasoning sequence $C$ conditioned on the current observation and the optional spatial prior.
    
    \item \textbf{Act (Sec.~\ref{subsec:act}):} A downstream action head consumes the latest latent reasoning states $H_{\text{reasoning}}$ and produces continuous action chunks for control. To support responsive execution, the reasoning module and the action head operate asynchronously at different frequencies.
\end{itemize}

\subsection{The ``Guide'' Phase: Multimodal Spatial Priors}
\label{subsec:guide}

\noindent\textbf{Spatial Prior Interface.}
We introduce an optional spatial prior $P_{\text{spatial}}$ as an additional input interface for sparse human guidance. The role of $P_{\text{spatial}}$ is not to replace the language instruction $L$, but to provide complementary geometric constraints in image space when the task is ambiguous or when targeted correction is needed. In practice, $P_{\text{spatial}}$ is specified on the primary camera view and is consumed jointly with the visual observation and language instruction.

\noindent\textbf{Source and Timing of Spatial Priors.}
The guidance interface is source-agnostic: spatial priors may come from scripted simulator annotations, external perception models such as Grounding-DINO or Gemini, or direct user intervention. They can also be provided under two timing regimes. An \emph{up-front} prior is attached at the episode start as additional task context, while a \emph{mid-episode} prior is injected when a human user or failure detector observes mis-grounding, an incorrect affordance, or an undesired motion path. In our batch evaluation of ambiguous scenarios, spatial priors are generated from simulator state for controlled comparison; in interactive settings, the intervention trigger is provided by the human user. Automatic triggering based on confidence, uncertainty, or failure detection is left for future work.

\noindent\textbf{Spatial Formulations.}
Our framework supports three levels of spatial guidance:
\begin{itemize}
    \item \textbf{Affordance Guide ($P_{\text{point}}$):} A single 2D image coordinate $(x, y)$ indicating a task-relevant affordance location, most commonly a grasp point, contact point, or interaction anchor on the target object. This is the lightest-weight form of intervention and is useful for rapidly specifying where the robot should interact.

    \item \textbf{Box Guide ($P_{\text{bbox}}$):} A bounding box $(x_{\min}, y_{\min}, x_{\max}, y_{\max})$ specifying a target region. This is useful when object identity or coarse localization is the bottleneck, although boxes may still include nearby distractors and therefore provide less precise affordance-level supervision than points.

    \item \textbf{Trace Guide ($P_{\text{trace}}$):} An ordered sequence of 2D points
    \begin{equation}
        P_{\text{trace}} = \big[(x_1, y_1), (x_2, y_2), \dots, (x_m, y_m)\big],
    \end{equation}
    representing a coarse image-space path. This form of guidance is useful for expressing directional preferences, motion style, or obstacle-avoidance cues.
\end{itemize}

\noindent\textbf{Serialization and Tokenization.}
To integrate $P_{\text{spatial}}$ into the VLM backbone, we serialize spatial priors into the model's coordinate token space and concatenate them with the textual instruction. Qwen3-VL natively supports point- and box-based localization in relative coordinate space. We therefore encode $P_{\text{point}}$ and $P_{\text{bbox}}$ using the same coordinate representation as the backbone's native grounding interface. For trace guidance, we represent the path as a short ordered sequence of point coordinates, using the same coordinate tokenization scheme for each waypoint.

\noindent\textbf{Fusion with the Observation Stream.}
After serialization, the spatial prior is provided together with the language instruction and visual observation, allowing the VLM to jointly attend to semantic content and human-provided geometric cues. This design preserves a unified inference interface: when $P_{\text{spatial}}$ is absent, the model operates fully autonomously; when it is present, the same backbone conditions its subsequent reasoning on the provided spatial prior without requiring a separate interaction-specific branch.

\subsection{The ``Think'' Phase: Conditioned Spatial-Visual CoT}
\label{subsec:think}

\noindent\textbf{Structured Reasoning Sequence.}
Given the augmented input tuple $(\mathcal{I}_t, L, s_t, P_{\text{spatial}})$, our model does not directly map observations to motor actions. Instead, the VLM first generates a structured spatial-visual reasoning sequence $C$ in an autoregressive manner. For both exposition and supervision, we organize this sequence into three functional segments,
\begin{equation}
    C = [C_{\text{task}}, C_{\text{vision}}, C_{\text{robot}}],
\end{equation}
which correspond to task decomposition, visual grounding, and robot-oriented motion reasoning, respectively.

\noindent\textbf{Reasoning Decomposition.}
\begin{enumerate}
    \item \textbf{Task CoT ($C_{\text{task}}$):} The model first produces a high-level semantic rationale that decomposes the instruction $L$ into executable sub-tasks and identifies the relevant objects and interactions required for completion.
    
    \item \textbf{Vision CoT ($C_{\text{vision}}$):} Conditioned on the observation and the preceding task rationale, the model predicts visually grounded intermediate targets, including target regions and task-relevant affordance locations in image space. This step anchors the semantic plan to concrete visual entities.
    
    \item \textbf{Robot CoT ($C_{\text{robot}}$):} Based on the grounded visual targets, the model further predicts a coarse image-space motion sketch for the end-effector, represented as a sequence of 2D waypoints that summarize the intended manipulation trajectory.
\end{enumerate}

\noindent\textbf{Conditioning on Human Spatial Guidance.}
The key property of this phase is that the reasoning process is explicitly conditioned on the optional spatial prior:
\begin{equation}
    P(C \mid \mathcal{I}_t, L, P_{\text{spatial}}).
\end{equation}
When $P_{\text{spatial}}$ is absent, the model reasons autonomously from the visual observation and language instruction alone. When a spatial prior is provided, it serves as an additional geometric constraint on the reasoning process. In particular, an affordance guide or box guide can anchor the model's visual grounding to a user-specified interaction point or region, while a trace guide can bias the predicted motion sketch toward a desired path. As a result, the same reasoning backbone supports both fully autonomous execution and interaction-conditioned correction under spatial ambiguity.

\noindent\textbf{Latent Reasoning States.}
Let $H_{\text{reasoning}}$ denote the hidden states associated with the generated reasoning tokens in $C$. These states provide a dense latent representation of the model's task understanding, visual grounding, and motion intent, and are passed to the downstream action head in the subsequent \textit{Act} phase.

\subsection{The ``Act'' Phase: Asynchronous Flow-Matching}
\label{subsec:act}

\noindent\textbf{Motivation: Decoupling Slow Reasoning from Fast Control.}
Autoregressive VLM reasoning is significantly slower than the control frequency typically required for closed-loop manipulation. If the model were forced to regenerate the full reasoning sequence $C$ at every control step, action execution would be limited by the decoding speed of the VLM, leading to delayed feedback and unstable behavior in dynamic interaction. To mitigate this mismatch, we separate high-level reasoning from low-level action generation.

\noindent\textbf{Asynchronous Execution Architecture.}
Our \textit{Act} phase adopts an asynchronous slow-fast design with two coupled modules:
\begin{itemize}
    \item \textbf{Slow Reasoning Module:} The VLM backbone executes the \textit{Guide} and \textit{Think} phases at a lower update frequency. Given the current multimodal input, it produces the structured reasoning sequence $C$ and the corresponding latent reasoning states
    \begin{equation}
        H_{\text{reasoning}} \in \mathbb{R}^{N \times D},
    \end{equation}
    where $N$ is the number of reasoning tokens and $D$ is the hidden dimension. These states summarize the current task decomposition, visual grounding, and motion intent, and are stored as the latest cached reasoning memory.
    
    \item \textbf{Fast Action Module:} A downstream Flow-Matching action head operates at a higher control frequency. At each control step, it receives the current observation together with the latest cached reasoning states $H_{\text{reasoning}}^{\text{latest}}$, and predicts continuous action chunks conditioned on this reasoning context. In our implementation, the action head accesses $H_{\text{reasoning}}^{\text{latest}}$ through cross-attention, allowing the control module to reuse the most recent reasoning output between VLM updates.
\end{itemize}

\noindent\textbf{Flow-Matching Action Generation.}
We parameterize action generation with a flow-matching policy head~\cite{black2024pi0,nvidia2025gr00t,zheng2025xvla}, which models continuous action chunks by learning a time-dependent vector field over actions. In our asynchronous design, the VLM and the action head consume different input streams at different update frequencies.

At a lower frequency, the VLM takes the primary image, the language instruction, and the optional spatial prior as input, i.e.,
\begin{equation}
    (\mathcal{I}^{\text{main}}_t, L, P_{\text{spatial}}) \;\longrightarrow\; C,\; H_{\text{reasoning}}.
\end{equation}
This stage produces the structured reasoning sequence $C$ and its corresponding latent reasoning states $H_{\text{reasoning}}$.

At a higher control frequency, the action head consumes the current control observation together with the latest cached reasoning states. Specifically, it takes the primary image, the wrist image, the proprioceptive state, and the most recent reasoning memory as input, and predicts
\begin{equation}
    v_\theta(x, \tau \mid \mathcal{I}^{\text{main}}_t, \mathcal{I}^{\text{wrist}}_t, s_t, H_{\text{reasoning}}^{\text{latest}}),
\end{equation}
where $x$ denotes the action variable and $\tau$ denotes the flow time. Integrating this vector field yields a continuous action chunk conditioned on both the current control observation and the latest available reasoning context.

This separation allows semantic and spatial reasoning to be updated at a lower frequency, while the action head continues to generate responsive actions at a higher frequency using the latest cached reasoning states. In this way, the policy preserves rich reasoning capacity without requiring autoregressive VLM decoding at every control step.
\subsection{Data Construction and Training Recipe}
\label{subsec:data_training}

\begin{figure}[t]
    \centering
    \includegraphics[width=1\linewidth]{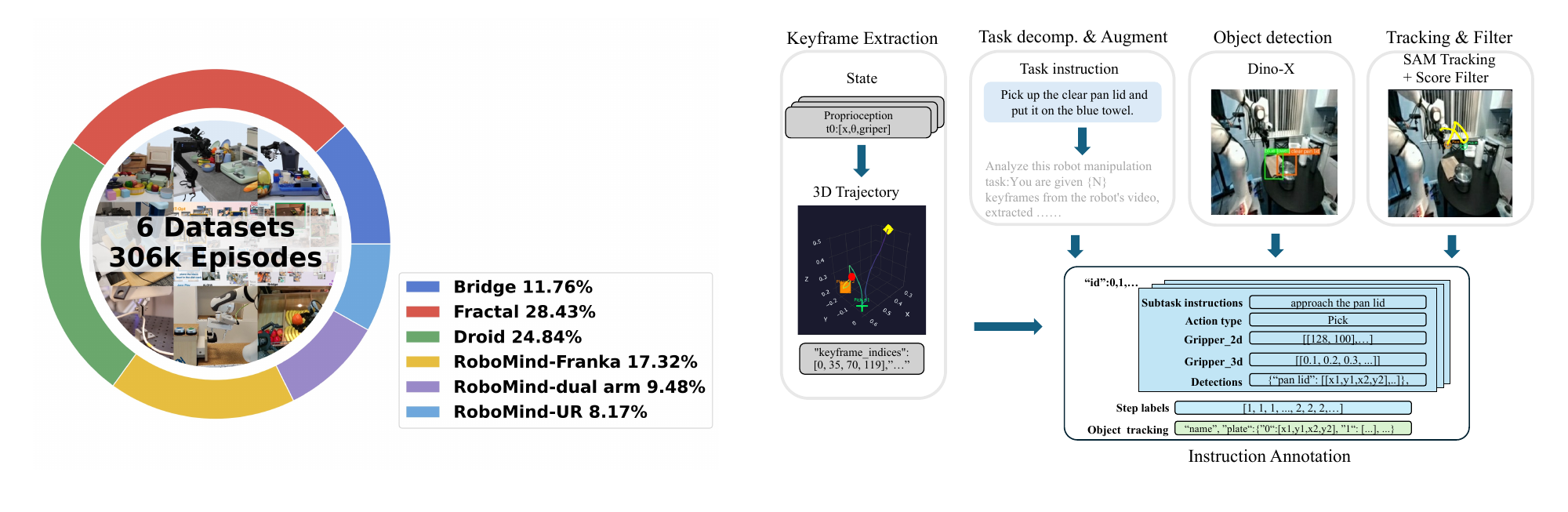}
    \caption{\textbf{Interact-306K and automatic instruction annotation.}
Left: Dataset composition: 306K episodes collected from six manipulation sources (\textit{e.g.}, Bridge~\cite{walke2023bridgedata}, Fractal~\cite{rt22023arxiv}, Droid~\cite{khazatsky2024droid}, and RoboMind variants~\cite{wu2025robomind}).
Right: Automatic annotation pipeline: keyframe extraction and task decomposition from trajectories, followed by open-vocabulary grounding and tracking to produce structured subtask instructions with temporally consistent object annotations.}
    \label{fig:data_engine}
\end{figure}

To train the framework at scale, we construct \textbf{Interact-306K}, a multi-embodiment dataset for guided spatial reasoning. As shown in Fig.\ref{fig:data_engine}, it is built from approximately 306K real-world manipulation trajectories collected from Open X-Embodiment (OXE)~\cite{o2024open}, DROID~\cite{khazatsky2024droid}, RoboMind~\cite{wu2025robomind}, and our own data, and augmented with automatically generated spatial-reasoning annotations.

\noindent\textbf{Automated Spatial-CoT Supervision.}
Since raw robot demonstrations do not contain explicit reasoning traces, we automatically construct supervision for both the \textit{Guide} and \textit{Think} phases. For each trajectory, we generate a structured reasoning target
\begin{equation}
    C = [C_{\text{task}}, C_{\text{vision}}, C_{\text{robot}}],
\end{equation}
aligned with the three-part decomposition used by our model: task decomposition is inferred from keyframes and language, visual grounding is obtained by localizing and tracking task-relevant objects in image space, and robot-centric supervision is derived by projecting end-effector motion into the primary view to produce affordance locations and coarse 2D motion sketches. To better match inference-time interaction, we further perturb the generated spatial annotations with stochastic noise, producing synthetic affordance points, boxes, and traces for training the \textit{Guide} interface.

\noindent\textbf{Training Recipe.}
We train the model in two stages. In Stage 1, we train the VLM backbone on Interact-306K to learn the \textit{Guide} and \textit{Think} components, using stochastic spatial conditioning so that the model is exposed to both guided and unguided inputs. We then train the Flow-Matching action head to map the latent reasoning states $H_{\text{reasoning}}$ together with control observations to continuous action chunks. In Stage 2, we jointly fine-tune the full policy on domain-specific robot data (e.g., BridgeData V2~\cite{walke2023bridgedata}) to adapt the reasoning module and action head to the target embodiment and environment. Unless otherwise specified, reasoning generation is optimized with autoregressive token prediction on $C$, while the action module is optimized with the standard flow-matching objective on action chunks.

\section{Experiments}
\label{sec:experiments}

We evaluate our method along three main axes: standard benchmark performance, out-of-distribution (OOD) robustness, and the effectiveness of explicit visual guidance under spatial ambiguity. We first assess autonomous performance on established manipulation benchmarks, including LIBERO~\cite{liu2023libero} and SimplerEnv~\cite{li24simpler}. We then evaluate OOD generalization using our proposed SimplerEnv-Plus benchmark, which introduces systematic perturbations across visual, robot, language, and object-centric factors. Finally, we study whether sparse visual guidance, such as affordance points and boxes, can effectively resolve ambiguity when language alone is insufficient.

\subsection{Experimental Setup}
\label{subsec:exp_setup}
We mainly evaluate our method on two simulation benchmarks: sim-to-sim Libero~\cite{liu2023libero} and real-to-sim: SimplerEnv~\cite{li24simpler}.

\noindent\textbf{Standard Benchmark for In-Domain Evaluation.}
We primarily evaluate our method on two simulation benchmarks: LIBERO~\cite{liu2023libero} and SimplerEnv~\cite{li24simpler}. LIBERO is a sim-to-sim benchmark covering diverse multi-task manipulation suites, including Spatial, Object, Goal, and Long, and is commonly used to evaluate multi-task generalization, instruction following, and long-horizon reasoning. SimplerEnv is a real-to-sim benchmark built on high-fidelity digital twins of real robot setups, and is designed to assess zero-shot visuomotor transfer and manipulation robustness under realistic visual conditions. In the main paper, we focus on the WidowX domain of SimplerEnv and defer additional results on Google Robot to the appendix.
\begin{table}[t]
    \centering
    \caption{\textbf{Main Results.} Success rates (\%) on the LIBERO and SimplerEnv (Bridge) benchmarks. * denotes reproduced results evaluated with a maximum inference horizon of 120 steps, consistent with the setting used for other models.}
    \label{tab:main}
    \resizebox{\linewidth}{!}{
    \begin{tabular}{l|cccc|c|cccc|c}
    \hline
    \multirow{2}{*}{Method} & \multicolumn{5}{c|}{LIBERO} & \multicolumn{5}{c}{SIMPLER-Env (Bridge)} \\
    \cline{2-11}
     & Spatial & Object & Goal & Long & Avg & Spoon & Carrot & Cube & Eggplant & Avg \\
    \hline
    OpenVLA~\cite{kimOpenVLAOpenSourceVisionLanguageAction2024}  & 84.7 & 88.4 & 79.2 & 53.7 & 76.5 & 4.2 & 0.0 & 8.3 & 45.8 & 14.6 \\
    OpenVLA-OFT~\cite{kim2025fine} & 96.2& 98.3& 96.2& 90.7& 95.3& - & - & - & - & - \\
    $\pi_0$~\cite{black2024pi0}  & 96.8 & 98.8 & 95.8 & 85.2 & 94.1 & 50.0 & 41.7 & 29.2 & 70.8 & 47.9 \\
    GR00T-N1~\cite{nvidia2025gr00t} & 94.4 & 97.6 & 93.0 & 90.6 & 93.9 & 64.5 & 65.5 & 5.5 & 93.0 & 57.1 \\
    $\pi_{0.5}$~\cite{blackP05VisionLanguageActionModel} & 98.8 & 98.2 & 98.0 & 92.4 & 96.9& - & - & - & - & - \\
    X-VLA*~\cite{zheng2025xvla}& 98.2 & 98.6 & 97.8 & \textbf{97.6} & 98.1& \textbf{95.8} & 75.0 & 62.5 & 70.8 & 76.0 \\
    ThinkAct~\cite{huangThinkActVisionLanguageActionReasoning2025} & 88.3 & 91.4 & 87.1 & 70.9 & 84.4 & 37.5 & 8.7 & 58.3 & 70.8 & 43.8 \\
    CoT-VLA~\cite{zhao2025cot}  & 87.5 & 91.6 & 87.6 & 69.0 & 81.1 & - & - & - & - & - \\
    Uni-VLA~\cite{wang2025unified} & 97.0 & \textbf{99.0} & 92.6 & 90.8 & 94.8 & 83.3 & 66.7 & 33.3 & \textbf{95.8} & 69.8\\
    MolmoAct~\cite{leeMolmoActActionReasoning2025} & 87.0 & 95.4 & 87.6 & 77.2 & 86.6 & - & - & - & - & - \\ 
    \hline
    GTA-VLA     &\textbf{99.0} & 98.8& \textbf{98.4} & \textbf{97.6}& \textbf{98.6}& \textbf{95.8} & \textbf{87.5} & \textbf{66.7} & 75.0 & \textbf{81.2} \\
    \hline
    \end{tabular}
    }
\end{table}
    
\noindent\textbf{SimplerEnv-Plus for OOD Evaluation.}
To evaluate robustness under systematic shift, we introduce \textbf{SimplerEnv-Plus}, an extended benchmark built on top of SimplerEnv. It includes four categories of perturbations:
\begin{itemize}
    \item \textit{Visual Shift:} We perturb low-level visual conditions through lighting variation and sensor viewpoint changes to test robustness to appearance changes.
    \item \textit{Robot State Shift:} We randomize the robot's initial state and introduce execution noise to simulate uncertainty in embodiment state and control.
    \item \textit{Language Shift:} We modify task instructions through lexical variation in verbs, nouns, and attributes to evaluate robustness to instruction diversity.
    \item \textit{Object Shift:} We replace standard targets with novel objects and introduce distractors to test zero-shot object generalization and robustness under perceptual ambiguity.
\end{itemize}

\noindent\textbf{Visual Guidance Evaluation.}
To evaluate the effectiveness of explicit visual guidance under ambiguity, we consider two challenging settings:
\begin{itemize}
    \item \textit{Unseen Object Ambiguity:} We replace standard training objects with novel instances from multiple categories, including \textit{Unseen Toy}, \textit{Unseen Fruit}, and \textit{Unseen Tool}. We then compare unguided execution against point- and box-guided execution to measure whether spatial priors improve zero-shot object grounding in unseen scenarios.
    
    \item \textit{Distractor-based Ambiguity:} We introduce same-category distractors that create fine-grained spatial ambiguity, including \textit{Color Distractors} (same category, different colors) and \textit{Position Distractors} (same object type at different locations). We compare unguided execution against point- and box-guided execution to evaluate whether visual guidance can resolve ambiguity when language alone is insufficient to uniquely specify the target.
\end{itemize}
\subsection{Main Results: Standard and OOD Performance}

\noindent\textbf{In-Domain Performance.}
As shown in Table~\ref{tab:main}, our method achieves strong in-domain performance on both LIBERO and SimplerEnv. On the highly competitive LIBERO benchmark, our approach reaches an average success rate of 98.6\%, performing on par with or slightly above the strongest baselines. This indicates that introducing explicit spatial reasoning does not compromise the policy's core manipulation ability or multi-task execution performance.

On the real-to-sim SimplerEnv benchmark, our method achieves an average success rate of 81.2\%, outperforming the strongest reported baseline. This gain is more pronounced than on LIBERO, suggesting that explicit spatial reasoning is particularly beneficial when policies must bridge the visual and semantic gap between open-world training data and simulated robot execution. By explicitly grounding task-relevant regions and affordances before action generation, the policy is better able to align semantic understanding with executable control.

\noindent\textbf{Out-of-Distribution Generalization.}
Table~\ref{tab:ood} shows that our method also improves robustness under systematic distribution shifts in SimplerEnv-Plus. Compared with baseline methods, our approach consistently maintains stronger performance across visual, robot-state, language, and object-centric perturbations.

Under \textit{Visual Shift}, our model remains more robust to sensor viewpoint changes and lighting variation, indicating that explicit intermediate grounding reduces reliance on spurious low-level correlations. Under \textit{Robot State Shift}, our method maintains strong performance despite randomized initial states and execution noise, suggesting that the combination of explicit reasoning and stable action generation improves robustness to embodiment uncertainty. Under \textit{Language Shift}, our model preserves competitive performance under lexical variation, showing that introducing spatial supervision does not substantially weaken language understanding. Finally, under \textit{Object Shift}, which includes both unseen objects and distractor-heavy scenes, our method shows the largest relative advantage, indicating that explicit task decomposition and grounded intermediate reasoning are especially helpful when target identification becomes ambiguous.

Overall, these results suggest that the proposed \textit{Think} and \textit{Act} design improves both in-domain execution and robustness under distribution shift, while preserving the ability to operate autonomously without external guidance.
\begin{table}[t]
    \centering
    \caption{\textbf{OOD Generalization on SimplerEnv-Plus.} Success rates (\%) under distribution shifts across visual, robot-state, language, and object-centric factors.}
    \label{tab:ood}
    \resizebox{\linewidth}{!}{
    \begin{tabular}{l|cc|cc|cc|c}
    \hline
    \multirow{2}{*}{Method} & \multicolumn{2}{c|}{Visual} & \multicolumn{1}{c|}{Robot} & \multicolumn{1}{c|}{Language} & \multicolumn{2}{c|}{Objects} & \multirow{2}{*}{Avg} \\
    \cline{2-7}
    & Sensor & Lighting & \multicolumn{1}{c|}{State} & \multicolumn{1}{c|}{Diversity} & Unseen Obj. & Distractor & \\
    
    \hline
    OpenVLA~\cite{kimOpenVLAOpenSourceVisionLanguageAction2024} & 5.2 & 6.3 & 0.0 & 8.3 & 2.1 & 0.0 & 3.7\\
    $\pi_{0.5}$~\cite{blackP05VisionLanguageActionModel} & 9.4 & 10.4 & 9.4 & 8.3 & 6.3 & 0.0 & 7.3\\
    X-VLA~\cite{zheng2025xvla} & 27.1 & 68.8 & 68.7 & 66.3 & 36.2 & 46.9 & 52.3\\
    \hline
    GTA-VLA & 39.6 & 76.1 & 79.2 & 68.1 &  58.3 &  50.0 & 61.4\\
    \hline
    \end{tabular}
    }
    \end{table}
\subsection{Guidance Efficacy}

Table~\ref{tab:guidance_efficacy} shows that explicit visual guidance consistently improves performance under both unseen-object ambiguity and distractor-based ambiguity. When language alone is insufficient to uniquely specify the correct target, sparse spatial priors provide a much stronger grounding signal than dense instruction. The two guidance forms show complementary behavior. Point guidance provides precise affordance-level supervision and is especially effective when same-category distractors create fine-grained spatial ambiguity. Box guidance provides stronger object-level localization and is more helpful when object identity is the main bottleneck, as in unseen-object settings. Overall, these results show that visual guidance is an effective mechanism for resolving ambiguity while allowing different spatial priors to target different failure modes.

\begin{table}[t]
    \centering
    \caption{Effectiveness of Visual Guidance in Ambiguous Scenarios. We evaluate different input modalities for our model on challenging SimplerEnv-Plus tasks. All values are success rates (\%). Visual guidance significantly outperforms even dense linguistic instructions, especially when spatial ambiguity is high.}
    \label{tab:guidance_efficacy}
    \resizebox{\columnwidth}{!}{%
    \begin{tabular}{l|ccc|c||cc|c}
    \toprule
    \multirow{2}{*}{\textbf{Guidance Modality}} & \multicolumn{4}{c|}{\textbf{Unseen Object Ambiguity}} & \multicolumn{3}{c}{\textbf{Distractor-based Ambiguity}} \\
    \cmidrule(r){2-5} \cmidrule(l){6-8}
     & Unseen Toy & Unseen Fruit & Unseen Tool & \textbf{Avg.} & Color Distractor & Pos. Distractor & \textbf{Avg.} \\
    \midrule
    Dense Instruction ($\pi_{0.5}$) & 8.3 & 12.5 & 8.3 & 9.7 & 8.3 & 8.3 & 8.3\\
    Dense Instruction(GTA-VLA) & 12.5 & 41.6 & 29.2 & 27.8 & 45.8 & 29.2 &  37.5\\
    \midrule
    \rowcolor{gray!20}
    + Visual Point Guide & 33.3 & 47.9 & 41.6 & 40.9 & 58.3 & 50.0 & 54.2\\
    \rowcolor{gray!20}
    + Visual Box Guide & 54.1 & 70.8 & 45.8 &  56.9 & 41.6 & 45.8 & 43.7\\
    \bottomrule
    \end{tabular}
    }
    \end{table}

\noindent\textbf{Trace guidance.}
Trace guidance targets motion preference, such as path shape and obstacle avoidance, rather than target selection. We therefore isolate it on obstacle-avoidance and constrained-path tasks, where the trace is constructed by linearly interpolating waypoints from the source gripper pose to the target box center. On this setting, trace guidance improves success from 27.8\% to 30.5\%, indicating that path-level priors can provide additional control beyond point- or box-level target disambiguation.

\subsection{Ablation Study}

Table~\ref{tab:ablation} ablates the three fields in the structured Chain-of-Thought on SimplerEnv-Bridge. Removing $C_{\text{vision}}$ causes the largest drop, showing that explicit visual grounding is central for cluttered scenes where target localization is the main bottleneck. Removing $C_{\text{task}}$ also substantially hurts performance, confirming that semantic decomposition is important for mapping instructions to objects and interactions. Removing $C_{\text{robot}}$ has a smaller but still measurable effect, suggesting that the action head can recover part of the low-level motion intent while still benefiting from explicit robot-oriented motion sketches.

\begin{table}[t]
    \centering
    \small
    \setlength{\tabcolsep}{3pt}
    \caption{Ablation of structured CoT fields and free-form CoT on SimplerEnv-Bridge. All values are success rates (\%).}
    \label{tab:ablation}
    \begin{tabular}{l|cccc|cc}
    \toprule
    \textbf{Setting} & \textbf{Spoon} & \textbf{Carrot} & \textbf{Cube} & \textbf{Eggplant} & \textbf{Avg} & \textbf{$\Delta$} \\
    \midrule
    \textbf{GTA-VLA (full)} & 95.8 & 87.5 & 66.7 & 75.0 & \textbf{81.2} & -- \\
    \midrule
    $-\,C_{\text{robot}}$  & 95.8 & 87.5 & 54.2 & 75.0 & 78.1 & $-3.1$ \\
    $-\,C_{\text{task}}$   & 95.8 & 91.7 & 54.2 & 33.3 & 68.8 & $-12.4$ \\
    $-\,C_{\text{vision}}$ & 95.8 & 79.2 & 41.7 & 50.0 & 66.7 & $-14.5$ \\
    \midrule
    Free-form CoT & 100.0 & 91.7 & 50.0 & 20.8 & 65.6 & $-15.6$ \\
    \bottomrule
    \end{tabular}
\end{table}

The free-form CoT variant underperforms the structured format under the same pseudo-label supervision. This suggests that, for the current scenarios, explicitly structured fields provide a more controllable and stable way to align task decomposition, visual grounding, and motion intent with user-provided priors. More discussion of this trade-off is provided in the appendix.

\subsection{Real-World Robot Deployment}
\label{subsec:real_world}
The real-world setup is shown in Figure~\ref{fig:real_world}, with additional implementation details provided in the appendix. We evaluate the model on four real-world picking tasks defined along two axes: whether the target object is seen or unseen, and whether the scene contains a single target or multiple same-category candidates requiring disambiguation. Concretely, the tasks include: (1) a single seen target in clutter, (2) a single unseen target in clutter, (3) a referred target among multiple identical seen objects, and (4) a referred target among multiple identical unseen objects. As shown in Figure~\ref{fig:real_world}, our method succeeds not only in cluttered seen-object settings, but also in more challenging unseen-object and referring scenarios, indicating that the proposed guidance and reasoning mechanism transfers effectively to real-world spatial ambiguity.

\begin{figure}[t]
    \centering
    \begin{subfigure}[t]{0.25\linewidth}
        \centering
        \includegraphics[width=\linewidth]{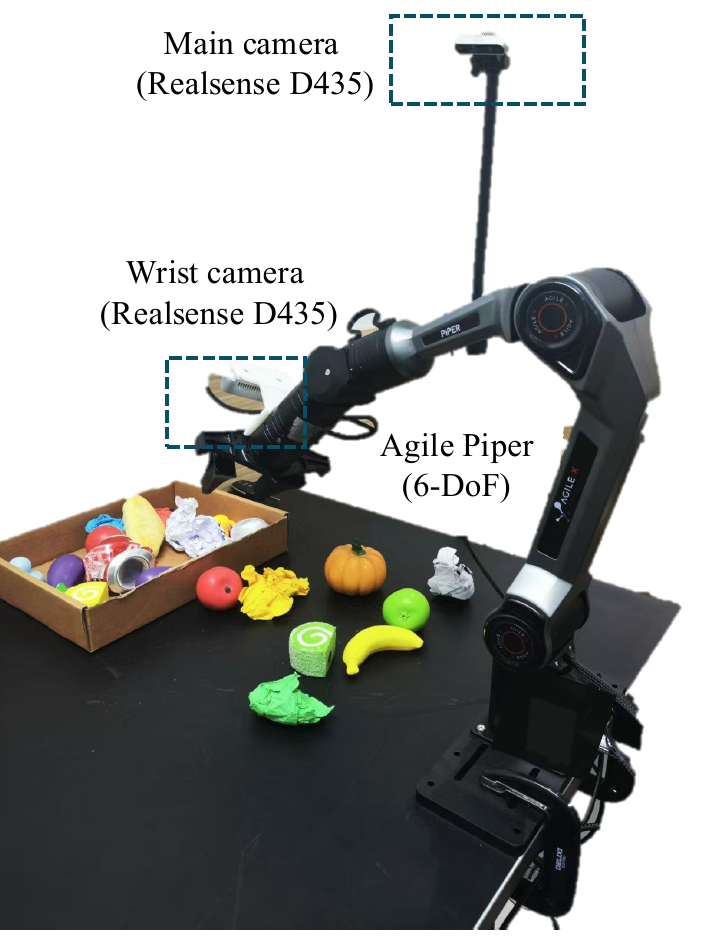}
        \label{fig:real_world_a}
    \end{subfigure}
    \hfill
    \begin{subfigure}[t]{0.73\linewidth}
        \centering
        \includegraphics[width=\linewidth]{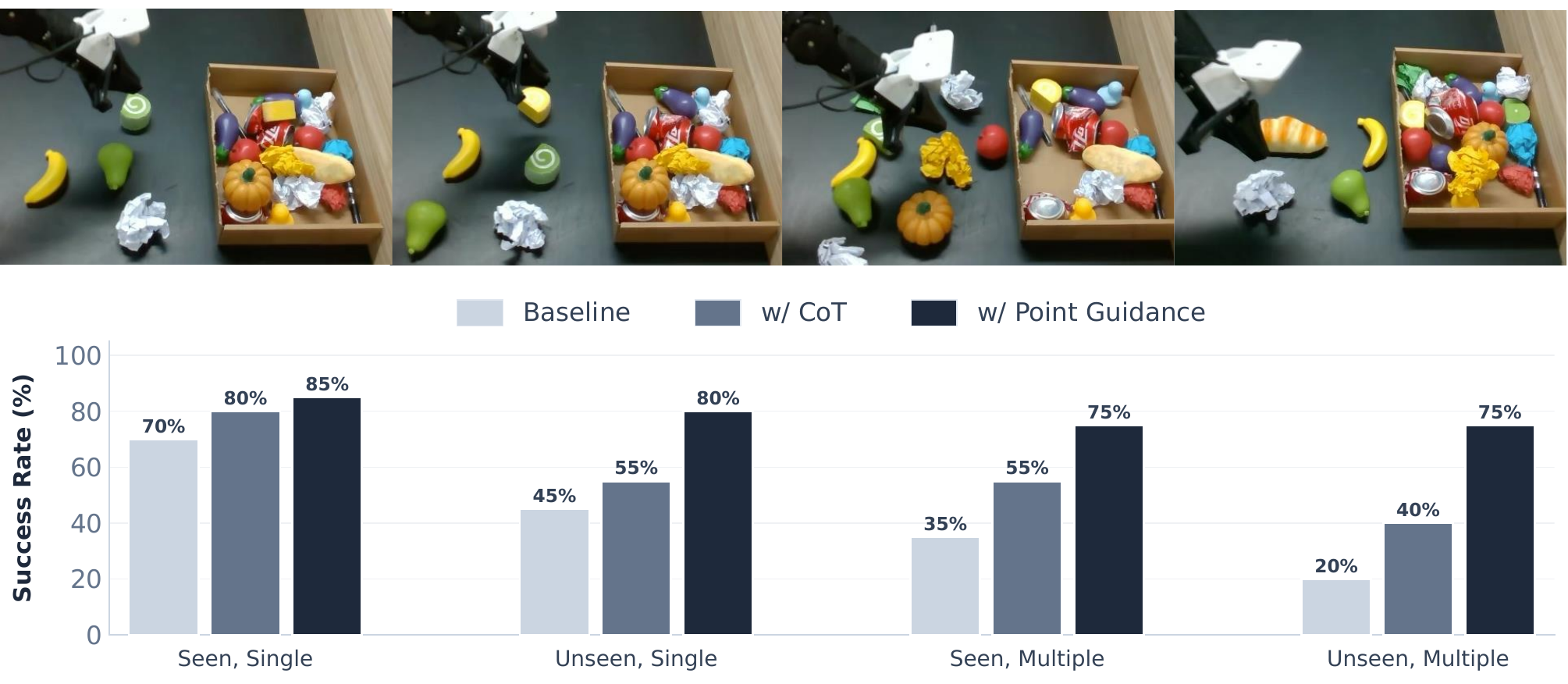}
        \label{fig:real_world_b}
    \end{subfigure}
    \caption{Real-world robot deployment. Left: the experimental setup with the Agile Piper robot, a primary, and a wrist-mounted camera. Right: qualitative examples and success rates across four picking tasks (seen/unseen targets and single/multiple candidate objects). Explicit reasoning improves over the baseline, and point guidance provides the largest gains in unseen and reference-ambiguous settings.}
    \label{fig:real_world}
\end{figure}

\section{Conclusion}

We presented \textbf{GTA-VLA} (\textbf{Guide, Think, Act}), an interactive Vision-Language-Action framework that enables spatially steerable embodied reasoning through  human visual guidance. By allowing spatial priors, such as affordance points, boxes, and traces, to directly condition a unified spatial-visual Chain-of-Thought, GTA-VLA moves beyond passive ``Sense-to-Act'' policies toward robot policies that are both autonomous and naturally correctable when failures or ambiguities arise. Experiments in both simulation and the real world show that our approach achieves strong autonomous performance while substantially improving failure recovery under out-of-domain shifts and spatial ambiguity. A current limitation of our framework is that both the reasoning process and the guidance interface are primarily formulated in 2D image space. An important direction for future work is to extend both the Chain-of-Thought representation and the visual guidance cues into 3D, enabling richer geometric grounding and more general interaction in real-world embodied settings.

\section*{Acknowledgements}
This work was partially supported by National Natural Science Foundation of China (Grant No. 62350710797).

%
\newpage
\bibliographystyle{splncs04}
\bibliography{main}

@String(CVPR  = {IEEE Conf. Comput. Vis. Pattern Recog.})

@String(ECCV  = {Eur. Conf. Comput. Vis.})

@String(NeurIPS = {Adv. Neural Inform. Process. Syst.})

@String(ICLR  = {Int. Conf. Learn. Represent.})

@String(CVPR  = {CVPR})

@String(ECCV  = {ECCV})

@String(NeurIPS = {NeurIPS})

@String(ICLR  = {ICLR})

@article{qwen3vl,                                                                                                                                                                      
    title={Qwen3-VL Technical Report},                                                                                                                                                 
    author={Bai, Shuai and Cai, Yuxuan and Chen, Ruizhe and Chen, Keqin and Chen, Xionghui and Cheng, Zesen and Deng, Lianghao and Ding, Wei and Gao, Chang and Ge, Chunjiang and Ge,    
  Wenbin and Guo, Zhifang and Huang, Qidong and Huang, Jie and Huang, Fei and Hui, Binyuan and Jiang, Shutong and Li, Zhaohai and Li, Mingsheng and Li, Mei and Li, Kaixin and Lin,      
  Zicheng and Lin, Junyang and Liu, Xuejing and Liu, Jiawei and Liu, Chenglong and Liu, Yang and Liu, Dayiheng and Liu, Shixuan and Lu, Dunjie and Luo, Ruilin and Lv, Chenxu and Men,
  Rui and Meng, Lingchen and Ren, Xuancheng and Ren, Xingzhang and Song, Sibo and Sun, Yuchong and Tang, Jun and Tu, Jianhong and Wan, Jianqiang and Wang, Peng and Wang, Pengfei and    
  Wang, Qiuyue and Wang, Yuxuan and Xie, Tianbao and Xu, Yiheng and Xu, Haiyang and Xu, Jin and Yang, Zhibo and Yang, Mingkun and Yang, Jianxin and Yang, An and Yu, Bowen and Zhang, Fei
   and Zhang, Hang and Zhang, Xi and Zheng, Bo and Zhong, Humen and Zhou, Jingren and Zhou, Fan and Zhou, Jing and Zhu, Yuanzhi and Zhu, Ke},
    journal={arXiv preprint arXiv:2511.21631},
    year={2025}
  }

@article{zheng2025xvla,
    title={X-VLA: Soft-Prompted Transformer as Scalable Cross-Embodiment Vision-Language-Action Model},
    author={Zheng, Jinliang and Li, Jianxiong and Wang, Zhihao and Liu, Dongxiu and Kang, Xirui and Feng, Yuchun and Zheng, Yinan and Zou, Jiayin and Chen, Yilun and Zeng, Jia and
  Zhang, Ya-Qin and Pang, Jiangmiao and Liu, Jingjing and Wang, Tai and Zhan, Xianyuan},
    journal={arXiv preprint arXiv:2510.10274},
    year={2025}
}

@article{nvidia2025gr00t,
    title={GR00T N1: An Open Foundation Model for Generalist Humanoid Robots},
    author={{NVIDIA} and Bjorck, Johan and Casta{\~n}eda, Fernando and Cherniadev, Nikita and Da, Xingye and Ding, Runyu and Fan, Linxi "Jim" and Fang, Yu and Fox, Dieter and Hu,
  Fengyuan and Huang, Spencer and Jang, Joel and Jiang, Zhenyu and Kautz, Jan and Kundalia, Kaushil and Lao, Lawrence and Li, Zhiqi and Lin, Zongyu and Lin, Kevin and Liu, Guilin and
  Llontop, Edith and Magne, Loic and Mandlekar, Ajay and Narayan, Avnish and Nasiriany, Soroush and Reed, Scott and Tan, You Liang and Wang, Guanzhi and Wang, Zu and Wang, Jing and
  Wang, Qi and Xiang, Jiannan and Xie, Yuqi and Xu, Yinzhen and Xu, Zhenjia and Ye, Seonghyeon and Yu, Zhiding and Zhang, Ao and Zhang, Hao and Zhao, Yizhou and Zheng, Ruijie and Zhu,
  Yuke},
    journal={arXiv preprint arXiv:2503.14734},
    year={2025}
  }

@article{huangFastThinkActEfficientVisionLanguageAction2026,
  title={Fast-ThinkAct: Efficient Vision-Language-Action Reasoning via Verbalizable Latent Planning},
  author={Huang, Chi-Pin and Man, Yunze and Yu, Zhiding and Chen, Min-Hung and Kautz, Jan and Wang, Yu-Chiang Frank and Yang, Fu-En},
  journal={arXiv preprint arXiv:2601.09708},
  year={2026}
}

@article{leeMolmoActActionReasoning2025,
  title={MolmoAct: Action Reasoning Models That Can Reason in Space},
  author={Lee, Jason and Duan, Jiafei and Fang, Haoquan and Deng, Yuquan and Liu, Shuo and Li, Boyang and Fang, Bohan and Zhang, Jieyu and Wang, Yi Ru and Lee, Sangho and Han, Winson and Pumacay, Wilbert and Wu, Angelica and Hendrix, Rose and Farley, Karen and VanderBilt, Eli and Farhadi, Ali and Fox, Dieter and Krishna, Ranjay},
  journal={arXiv preprint arXiv:2508.07917},
  year={2025}
}

@article{seed2025seed1_5vl,
  title={Seed1.5-VL Technical Report},
  author={ByteDance Seed Team},
  journal={arXiv preprint arXiv:2505.07062},
  year={2025}
}

@article{li24simpler,
         title={Evaluating Real-World Robot Manipulation Policies in Simulation},
         author={Xuanlin Li and Kyle Hsu and Jiayuan Gu and Karl Pertsch and Oier Mees and Homer Rich Walke and Chuyuan Fu and Ishikaa Lunawat and Isabel Sieh and Sean Kirmani and Sergey Levine and Jiajun Wu and Chelsea Finn and Hao Su and Quan Vuong and Ted Xiao},
         journal = {arXiv preprint arXiv:2405.05941},
         year={2024}
}

@article{tangMindHandPurposeful2025,
  title={Mind to Hand: Purposeful Robotic Control via Embodied Reasoning},
  author={Tang, Peijun and Xie, Shangjin and Sun, Binyan and Huang, Baifu and Luo, Kuncheng and Yang, Haotian and Jin, Weiqi and Wang, Jianan},
  journal={arXiv preprint arXiv:2512.08580},
  year={2025}
}

@inproceedings{zawalskiRoboticControlEmbodied2025a,
  title={Robotic Control via Embodied Chain-of-Thought Reasoning},
  author={Zawalski, Micha{\l} and Chen, William and Pertsch, Karl and Mees, Oier and Finn, Chelsea and Levine, Sergey},
  booktitle={CoRL},
  pages={3157--3181},
  year={2025},
  organization={PMLR}
}

@article{black2024pi0,
  title={{$\pi_0$}: A Vision-Language-Action Flow Model for General Robot Control},
  author={Black, Kevin and Brown, Noah and Driess, Danny and Esmail, Adnan and Equi, Michael and Finn, Chelsea and Fusai, Niccolo and Groom, Lachy and Hausman, Karol and Ichter, Brian and others},
  journal={arXiv preprint arXiv:2410.24164},
  year={2024}
}

@article{blackP05VisionLanguageActionModel,
  title={{$\pi_{0.5}$}: A Vision-Language-Action Model with Open-World Generalization},
  author={Black, Kevin and Brown, Noah and Driess, Danny and Esmail, Adnan and Equi, Michael and Finn, Chelsea and Fusai, Niccolo and others},
  journal={arXiv preprint arXiv:2504.16054},
  year={2025}
}

@article{cheangGR3TechnicalReport2025,
  title={GR-3 Technical Report},
  author={Cheang, Chilam and Chen, Sijin and Cui, Zhongren and Hu, Yingdong and Huang, Liqun and Kong, Tao and Li, Hang and Li, Yifeng and Liu, Yuxiao and Ma, Xiao and Niu, Hao and Ou, Wenxuan and Peng, Wanli and Ren, Zeyu and Shi, Haixin and Tian, Jiawen and Wu, Hongtao and Xiao, Xin and Xiao, Yuyang and Xu, Jiafeng and Yang, Yichu},
  journal={arXiv preprint arXiv:2507.15493},
  year={2025}
}

@article{chenInternVLAM1SpatiallyGuided2025,
  title={InternVLA-M1: A Spatially Guided Vision-Language-Action Framework for Generalist Robot Policy},
  author={Chen, Xinyi and Chen, Yilun and Fu, Yanwei and Gao, Ning and Jia, Jiaya and Jin, Weiyang and Li, Hao and Mu, Yao and Pang, Jiangmiao and Qiao, Yu and Tian, Yang and Wang, Bin and Wang, Bolun and Wang, Fangjing and Wang, Hanqing and Wang, Tai and Wang, Ziqin and Wei, Xueyuan and Wu, Chao and Yang, Shuai and Ye, Jinhui and Yu, Junqiu and Zeng, Jia and Zhang, Jingjing and Zhang, Jinyu and Zhang, Shi and Zheng, Feng and Zhou, Bowen and Zhu, Yangkun},
  journal={arXiv preprint arXiv:2510.13778},
  year={2025}
}

@article{chiDiffusionPolicyVisuomotor2023,
  title={Diffusion policy: Visuomotor policy learning via action diffusion},
  author={Chi, Cheng and Xu, Zhenjia and Feng, Siyuan and Cousineau, Eric and Du, Yilun and Burchfiel, Benjamin and Tedrake, Russ and Song, Shuran},
  journal={The International Journal of Robotics Research},
  volume={44},
  number={10-11},
  pages={1684--1704},
  year={2025},
  publisher={Sage Publications Sage UK: London, England}
}

@inproceedings{dengGraspVLAGraspingFoundation2025,
  title={GraspVLA: A Grasping Foundation Model Pre-trained on Billion-scale Synthetic Action Data},
  author={Deng, Shengliang and Yan, Mi and Wei, Songlin and Ma, Haixin and Yang, Yuxin and Chen, Jiayi and Zhang, Zhiqi and Yang, Taoyu and Zhang, Xuheng and Cui, Heming and others},
  booktitle={CoRL},
  year={2025},
}

@inproceedings{kimOpenVLAOpenSourceVisionLanguageAction2024,
  title={OpenVLA: An Open-Source Vision-Language-Action Model},
  author={Kim, Moo Jin and Pertsch, Karl and Karamcheti, Siddharth and Xiao, Ted and Balakrishna, Ashwin and Nair, Suraj and Rafailov, Rafael and Foster, Ethan P and Sanketi, Pannag R and Vuong, Quan and others},
  booktitle={CoRL},
  year={2025},
}

@article{liCogACTFoundationalVisionLanguageAction2024,
  title={CogACT: A Foundational Vision-Language-Action Model for Synergizing Cognition and Action in Robotic Manipulation},
  author={Li, Qixiu and Liang, Yaobo and Wang, Zeyu and Luo, Lin and Chen, Xi and Liao, Mozheng and Wei, Fangyun and Deng, Yu and Xu, Sicheng and Zhang, Yizhong and Wang, Xiaofan and Liu, Bei and Fu, Jianlong and Bao, Jianmin and Chen, Dong and Shi, Yuanchun and Yang, Jiaolong and Guo, Baining},
  journal={arXiv preprint arXiv:2411.19650},
  year={2024}
}

@article{teamOctoOpenSourceGeneralist2024,
  title={Octo: An Open-Source Generalist Robot Policy},
  author={Team, Octo Model and Ghosh, Dibya and Walke, Homer and Pertsch, Karl and Black, Kevin and Mees, Oier and Dasari, Sudeep and Hejna, Joey and Kreiman, Tobias and Xu, Charles and Luo, Jianlan and Tan, You Liang and Chen, Lawrence Yunliang and Sanketi, Pannag and Vuong, Quan and Xiao, Ted and Sadigh, Dorsa and Finn, Chelsea and Levine, Sergey},
  journal={arXiv preprint arXiv:2405.12213},
  year={2024}
}

@inproceedings{o2024open,
  title={Open x-embodiment: Robotic learning datasets and rt-x models: Open x-embodiment collaboration 0},
  author={O’Neill, Abby and Rehman, Abdul and Maddukuri, Abhiram and Gupta, Abhishek and Padalkar, Abhishek and Lee, Abraham and Pooley, Acorn and Gupta, Agrim and Mandlekar, Ajay and Jain, Ajinkya and others},
  booktitle={IEEE International Conference on Robotics and Automation},
  year={2024},
}

@inproceedings{walke2023bridgedata,
    title={BridgeData V2: A Dataset for Robot Learning at Scale},
    author={Walke, Homer and Black, Kevin and Lee, Abraham and Kim, Moo Jin and Du, Max and Zheng, Chongyi and Zhao, Tony and Hansen-Estruch, Philippe and Vuong, Quan and He, Andre and Myers, Vivek and Fang, Kuan and Finn, Chelsea and Levine, Sergey},
    booktitle={CoRL},
    year={2023}
}

@article{ravi2024sam2,
  title={SAM 2: Segment Anything in Images and Videos},
  author={Ravi, Nikhila and Gabeur, Valentin and Hu, Yuan-Ting and Hu, Ronghang and Ryali, Chaitanya and Ma, Tengyu and Khedr, Haitham and R{\"a}dle, Roman and Rolland, Chloe and Gustafson, Laura and Mintun, Eric and Pan, Junting and Alwala, Kalyan Vasudev and Carion, Nicolas and Wu, Chao-Yuan and Girshick, Ross and Doll{\'a}r, Piotr and Feichtenhofer, Christoph},
  journal={arXiv preprint arXiv:2408.00714},
  year={2024}
}

@article{jiang2025detect,
  title={Detect anything via next point prediction},
  author={Jiang, Qing and Huo, Junan and Chen, Xingyu and Xiong, Yuda and Zeng, Zhaoyang and Chen, Yihao and Ren, Tianhe and Yu, Junzhi and Zhang, Lei},
  journal={arXiv preprint arXiv:2510.12798},
  year={2025}
}

@inproceedings{wu2025robomind,
  title={Robomind: Benchmark on multi-embodiment intelligence normative data for robot manipulation},
  author={Wu, Kun and Hou, Chengkai and Liu, Jiaming and Che, Zhengping and Ju, Xiaozhu and Yang, Zhuqin and Li, Meng and Zhao, Yinuo and Xu, Zhiyuan and Yang, Guang and others},
  booktitle={Robotics: Science and Systems}, 
  year={2025},
}

@inproceedings{khazatsky2024droid,
    title={DROID: A large-scale in-the-wild robot manipulation dataset},
  author={Khazatsky, Alexander and Pertsch, Karl and Nair, Suraj and Balakrishna, Ashwin and Dasari, Sudeep and Karamcheti, Siddharth and Nasiriany, Soroush and Srirama, Mohan Kumar and Chen, Lawrence Yunliang and Ellis, Kirsty and others},
  booktitle={Robotics: Science and Systems},
  year={2024}
}

@inproceedings{rt22023arxiv,
    title={Rt-2: Vision-language-action models transfer web knowledge to robotic control},
  author={Zitkovich, Brianna and Yu, Tianhe and Xu, Sichun and Xu, Peng and Xiao, Ted and Xia, Fei and Wu, Jialin and Wohlhart, Paul and Welker, Stefan and Wahid, Ayzaan and others},
  booktitle={CoRL},
  year={2023},
}

@inproceedings{kirillov2023segany,
  title={Segment anything},
  author={Kirillov, Alexander and Mintun, Eric and Ravi, Nikhila and Mao, Hanzi and Rolland, Chloe and Gustafson, Laura and Xiao, Tete and Whitehead, Spencer and Berg, Alexander C and Lo, Wan-Yen and others},
  booktitle={CVPR},
  year={2023}
}

@article{carionSAM3Segment2025,
  title={SAM 3: Segment Anything with Concepts},
  author={Carion, Nicolas and Gustafson, Laura and Hu, Yuan-Ting and Debnath, Shoubhik and Hu, Ronghang and Suris, Didac and Ryali, Chaitanya and Alwala, Kalyan Vasudev and Khedr, Haitham and Huang, Andrew and Lei, Jie and Ma, Tengyu and Guo, Baishan and Kalla, Arpit and Marks, Markus and Greer, Joseph and Wang, Meng and Sun, Peize and R{\"a}dle, Roman and Afouras, Triantafyllos and Mavroudi, Effrosyni and Xu, Katherine and Wu, Tsung-Han and Zhou, Yu and Momeni, Liliane and Hazra, Rishi and Ding, Shuangrui and Vaze, Sagar and Porcher, Francois and Li, Feng and Li, Siyuan and Kamath, Aishwarya and Cheng, Ho Kei and Doll{\'a}r, Piotr and Ravi, Nikhila and Saenko, Kate and Zhang, Pengchuan and Feichtenhofer, Christoph},
  journal={arXiv preprint arXiv:2511.16719},
  year={2025}
}

@inproceedings{Jiang2024TRex2TG,
  title={T-rex2: Towards generic object detection via text-visual prompt synergy},
  author={Jiang, Qing and Li, Feng and Zeng, Zhaoyang and Ren, Tianhe and Liu, Shilong and Zhang, Lei},
  booktitle={ECCV},
  year={2024},
}

@inproceedings{You2023FerretRA,
  title={Ferret: Refer and Ground Anything Anywhere at Any Granularity},
  author={You, Haoxuan and Zhang, Haotian and Gan, Zhe and Du, Xianzhi and Zhang, Bowen and Wang, Zirui and Cao, Liangliang and Chang, Shih-Fu and Yang, Yinfei},
  booktitle={ICLR},
  year={2023}
}

@article{Yang2023SetofMarkPU,
  title={Set-of-Mark Prompting Unleashes Extraordinary Visual Grounding in GPT-4V},
  author={Jianwei Yang and Hao Zhang and Feng Li and Xueyan Zou and Chun-yue Li and Jianfeng Gao},
  journal={arXiv preprint arXiv:2310.11441},
  year={2023}
}

@article{Chen2023ShikraUM,
  title={Shikra: Unleashing Multimodal LLM's Referential Dialogue Magic},
  author={Ke Chen and Zhao Zhang and Weili Zeng and Richong Zhang and Feng Zhu and Rui Zhao},
  journal={arXiv preprint arXiv:2306.15195},
  year={2023}
}

@article{Belkhale2024RTHAH,
  title={RT-H: Action Hierarchies Using Language},
  author={Suneel Belkhale and Tianli Ding and Ted Xiao and Pierre Sermanet and Quon Vuong and Jonathan Tompson and Yevgen Chebotar and Debidatta Dwibedi and Dorsa Sadigh},
  journal={arXiv preprint arXiv:2403.01823},
  year={2024}
}

@inproceedings{Gu2023RTTrajectoryRT,
  title={RT-Trajectory: Robotic Task Generalization via Hindsight Trajectory Sketches},
  author={Gu, Jiayuan and Kirmani, Sean and Wohlhart, Paul and Lu, Yao and Arenas, Montserrat Gonzalez and Rao, Kanishka and Yu, Wenhao and Fu, Chuyuan and Gopalakrishnan, Keerthana and Xu, Zhuo and others},
  year={2023},
  booktitle={ICLR}
}

@inproceedings{huangThinkActVisionLanguageActionReasoning2025,
  title={ThinkAct: Vision-Language-Action Reasoning via Reinforced Visual Latent Planning},
  author={Huang, Chi-Pin and Wu, Yueh-Hua and Chen, Min-Hung and Wang, Yu-Chiang Frank and Yang, Fu-En},
  booktitle={NeurIPS},
  year={2025}
}

@article{liu2023libero,
  title={Libero: Benchmarking knowledge transfer for lifelong robot learning},
  author={Liu, Bo and Zhu, Yifeng and Gao, Chongkai and Feng, Yihao and Liu, Qiang and Zhu, Yuke and Stone, Peter},
  journal={NeurIPS},
  volume={36},
  pages={44776--44791},
  year={2023}
}

@article{kim2025fine,
  title={Fine-tuning vision-language-action models: Optimizing speed and success},
  author={Kim, Moo Jin and Finn, Chelsea and Liang, Percy},
  journal={arXiv preprint arXiv:2502.19645},
  year={2025}
}

@inproceedings{zhao2025cot,
  title={Cot-vla: Visual chain-of-thought reasoning for vision-language-action models},
  author={Zhao, Qingqing and Lu, Yao and Kim, Moo Jin and Fu, Zipeng and Zhang, Zhuoyang and Wu, Yecheng and Li, Zhaoshuo and Ma, Qianli and Han, Song and Finn, Chelsea and others},
  booktitle={CVPR},
  pages={1702--1713},
  year={2025}
}

@article{wang2025unified,
  title={Unified vision-language-action model},
  author={Wang, Yuqi and Li, Xinghang and Wang, Wenxuan and Zhang, Junbo and Li, Yingyan and Chen, Yuntao and Wang, Xinlong and Zhang, Zhaoxiang},
  journal={arXiv preprint arXiv:2506.19850},
  year={2025}
}

\end{document}